# 3DTINC: Time-Equivariant Non-Contrastive Learning for Predicting Disease Progression from Longitudinal OCTs

Taha Emre, Arunava Chakravarty, Antoine Rivail, Dmitrii Lachinov, Oliver Leingang, Sophie Riedl, Julia Mai, Hendrik P.N. Scholl, Sobha Sivaprasad, Daniel Rueckert, Andrew Lotery, Ursula Schmidt-Erfurth, and Hrvoje Bogunović, for the PINNACLE Consortium

**Abstract**—Self-supervised learning (SSL) has emerged as a powerful technique for improving the efficiency and effectiveness of deep learning models. Contrastive methods are a prominent family of SSL that extract similar representations of two augmented views of an image while pushing away others in the representation space as negatives. However, the state-of-the-art contrastive methods require large batch sizes and augmentations designed for natural images that are impractical for 3D medical images. To address these limitations, we propose a new longitudinal SSL method, 3DTINC, based on non-contrastive learning. It is designed to learn perturbation-invariant features for 3D optical coherence tomography (OCT) volumes, using augmentations specifically designed for OCT. We introduce a new non-contrastive similarity loss term that learns temporal information implicitly from intra-patient scans acquired at different times. Our experiments show that this temporal information is crucial for predicting progression of retinal diseases, such as age-related macular degeneration (AMD). After pretraining with 3DTINC, we evaluated the learned representations and the prognostic models on two large-scale longitudinal datasets of retinal OCTs where we predict the conversion to wet-AMD within a six-month interval. Our results demonstrate that each component of our contributions is crucial for learning meaningful representations useful in predicting disease progression from longitudinal volumetric scans.

Manuscript submitted May, 06, 2023. This work was supported in part by Wellcome Trust Collaborative Award (PINNACLE) Ref. 210572/Z/18/Z, and Austrian Science Fund (FWF) [10.55776/FG9]. For the purpose of open access, the author has applied a CC-BY public copyright licence to any author accepted manuscript version arising from this submission.

Taha Emre, Arunava Chakravarty, Antoine Rivail, Dmitrii Lachinov, Oliver Leingang, Sophie Riedl, Julia Mai, Ursula Schmidt-Erfurth, and Hrvoje Bogunović are with the Department of Ophthalmology and Optometry, Medical University of Vienna, Austria. Hrvoje Bogunović is also with the Christian Doppler Lab for Artificial Intelligence in Retina, Medical University of Vienna, Austria (hrvoje.bogunovic@meduniwien.ac.at).

Sobha Sivaprasad is with the NIHR Moorfields Biomedical Research Centre, Moorfields Eye Hospital NHS Foundation Trust, London, United Kingdom.

Daniel Rueckert is with BioMedIA, Imperial College London, United Kingdom, and also with the Institute for AI and Informatics in Medicine, Klinikum rechts der Isar, Technical University Munich, Germany.

Hendrik P.N. Scholl is with Institute of Molecular and Clinical Ophthalmology Basel, and Department of Ophthalmology, University of Basel, Basel, Switzerland.

Andrew Lotery is with Clinical and Experimental Sciences, Faculty of Medicine, University of Southampton, United Kingdom

learning, disease progression, longitudinal imaging, optical coherence tomography, retina

## I. INTRODUCTION

Age-related macular degeneration (AMD) is the leading cause of deterioration and complete loss of central vision in the elderly population [1]. AMD progresses to either geographic atrophy (GA, also called dry-AMD), or neovascular AMD (nAMD, also called wet-AMD). Wet-AMD is characterized by the formation of new vessels in the choroid, which leak fluid into the tissue, causing scarring and degradation of the vision [2]. The activity of wet-AMD can be mitigated by a frequent, e.g. bimonthly, intravitreal injection of anti-VEGF agents, most effective if applied soon after the conversion event [3]. However, the conversion from intermediate to wet-AMD can appear suddenly, and its clinical indicators are still not well-understood (Fig. 1). Thus, due to a lack of knowledge on wet-AMD conversion biomarkers, it is currently difficult for ophthalmologists to accurately determine the risk of conversion. This is especially relevant for the *fellow-eye* of an already treated eye to prevent vision loss in both eyes.

Retinal diseases including AMD are commonly diagnosed using non-invasive optical coherence tomography (OCT) scans. They provide a dense 3D volume consisting of a series of 2D cross-sectional slices (B-scans) with a resolution of a few micrometers. OCT scans are especially valuable to guide the treatment of patients with wet-AMD, where scans of both eyes are typically acquired across multiple visits, creating a time series depicting the progression of AMD in the fellow eyes, which are often still in the intermediate stage. Such longitudinal imaging datasets are an invaluable repository for developing deep learning models that can be trained to detect imaging patterns of imminent disease conversion.

The volumetric and longitudinal nature of OCT datasets presents a special challenge for developing deep learning models, as they require large annotated volumetric datasets and large computational resources, e.g. GPU memory. Moreover, it has been shown that, when trained from randomly initialized weights, the benefit of rich 3D information is canceled out by the difficulty of training a 3D model with volumetric scans [4]. This calls for new deep learning approaches that allow exploiting such large sets of unlabeled data, can utilize temporal and 3D volumetric information, and are additionally memory efficient. Such pretrained models are expected to learn temporally-informed representations and to be efficiently fine-tuned for downstream prognostic tasks. Current contrastive methods are based on samples with the assumption of independence and identical distribution (i.i.d). However, in temporal datasets, this is often not the case, as there is a high degree of correlation between the OCT scans of the same patient acquired across multiple visits at varying time intervals, which should be reflected in the learned representational space. Furthermore, disease progression prediction is inherently a temporal task. Therefore, OCT scan representations should capture potential future evolutions to achieve good prediction performance. One way of doing this is by aligning the representations of a patient's scan with respect to a common trajectory. This alignment process requires a signal; in a supervised setting, this could be the change in fluid volume or other trackable biomarkers that progress over time. However, in an *unsupervised setting*, the only available and consistent option is the time difference, assuming that disease severity can only stagnate or progress, and cannot reverse. In summary, the goal is to learn a smooth trajectory from longitudinal scans implicitly. An ideal representation for prognosis should be sensitive to changes in OCT volumes due to the progression of time. In other words, the representations should be *equivariant* in time such that the change in representation space should be proportional to the time component in the OCT volume domain.

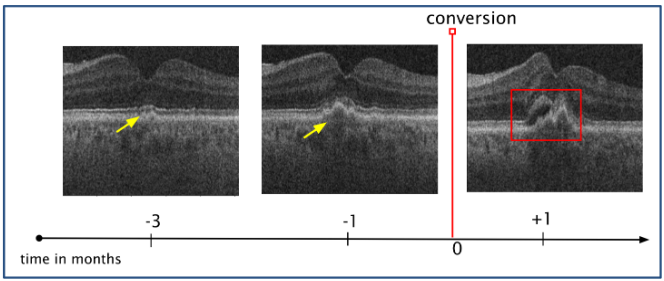

Fig. 1: An example of wet-AMD conversion. For the first two OCT scans, the eye is in the intermediate stage and only small changes are visible over the visits. On the last OCT scan, conversion to wet-AMD has already occurred, and there are drastic changes in the retinal structure.

In this paper, we present a deep learning method, 3D Temporally Informed Non-Contrastive Learning (**3DTINC**), that improves performance on limited downstream datasets by utilizing unlabeled longitudinal data in a self-supervised manner. We first set up an augmentation scheme for non-contrastive training suitable for 3D OCT volumes. Then, our novel non-contrastive loss term based on VICReg [5], forces the convolutional neural network (CNN) to produce representations which implicitly encode the temporal information from longitudinal data. Finally, we test our method on a prediction task of wet-AMD conversion, using fellow-eye retinal OCTs coming from multiple clinical datasets. Results demonstrate that the embedded temporal knowledge is crucial for time-relevant tasks such as AMD conversion prediction. To the best of our knowledge, this is the first work on 3D non-contrastive learning for OCT scans. It extends and advances on our prior work on 2D non-contrastive learning for AMD conversion prediction [6]. In summary, our contributions on top of our previous method [6] are:

- We process dense 3D OCT volumes with an efficient video CNN model
- We propose a novel augmentation scheme for non-contrastive pretraining specific to the 3D OCT volumes
- We analyze our novel loss function to show how it favours learning time-equivariant representations and also verify this property experimentally
- We test the extracted representations on an external dataset acquired with a different OCT scanner and show that our 3DTINC produces representations more robust to domain-shift

## II. BACKGROUND AND RELATED WORK

Self-supervised learning (SSL) is a pretraining strategy for learning meaningful representations of data in an unsupervised manner. SSL aims to learn data representations that capture important features specific to the input imaging modality. Initializing networks with the SSL-pretrained weights can improve their performance on the downstream task while reducing training data and time. It has been a common approach when the labels are scarce or noisy to stabilize the optimization by preventing over-fitting for the downstream task [7]. Usually, SSL pretraining involves solving a pretext task that does not require manual labeling. A model should learn useful representations for the downstream task by solving the pretext task successfully. Predicting relative image patch positions [8], jigsaw-puzzle solving [9] and image rotation degree prediction [10] are some of the most common pretext tasks. But these methods often learn representations specific to the heuristic pretext task and do not generalize well to different downstream tasks [11].

Contrastive learning emerged as a pretraining SSL step for extracting representations without relying on a pretext task. The core idea of contrastive methods is to learn image representations that are invariant to certain transformations, representative of irrelevant data variations. First, an image is augmented twice with the transformations, resulting in two *views*. Then, the network produces two representations. Finally, a contrastive loss, typically InfoNCE loss [12] is used to bring the two representations of the same image together in the embedding space, while it treats other image representations as negatives and pushes them apart. Contrastive methods can be interpreted as instance discrimination methods [13]. Ideally, the network should not produce the same representation for all the inputs while outputting very similar representations for the transformed versions of images. Sometimes, the network learns a shortcut for minimizing the distance by outputting the same representations for all samples: a trivial solution named *representational collapse*, which leads to severe degradation of the downstream task performance. Avoiding the collapse of representations requires the success of pulling and pushing operation, but it demands very large batches, with thousands of samples [11] and strong image transformations to prevent the network from learning shortcuts. Alternatively, another memory-intensive approach is to keep a dictionary of all representations and use it for dissimilar examples [14].

Non-contrastive methods [15], unlike contrastive ones [11], do not require large batch sizes, which is crucial for already memory-heavy 3D models. Additionally, they do not define different pairs within a batch as hard negatives, which is a desired property for medical imaging since different samples can share similar semantic information, such as similar anatomy or disease stage. Still, non-contrastive methods need to be able to distinguish different samples in the representational space. Therefore, instead of pushing different negative samples apart and pulling positive pairs together, non-contrastive methods learn to produce distinct and meaningful representations in several ways, e.g., by stopping gradients in one of the network branches to create discrepancy [15], predicting the output of the other branch [16], or prototype clustering [17]. In general, non-contrastive methods can be grouped into two categories. The first group like [15]–[17] creates a discrepancy between branches to avoid representational collapse. Another group of non-contrastive methods like VICReg [5] and Barlow Twins [18] employs redundancy reduction by defining appropriate loss functions that aim to increase correlation between the input pairs and to decorrelate the rest. They can be interpreted as maximizing the information in an embedding [19]. In this aspect, they prevent a second type of collapse called *informational collapse* [20], where the majority of the variables in a representation vector (embedding) does not contain information about the input. All the aforementioned strategies allow to learn meaningful representations without the need for hard negatives and prevent informational and representational collapse.

Most medical imaging datasets suffer from label scarcity, imbalance and noisiness. However, a large amount of unlabeled patient data is available and can be used for pretraining to improve the downstream task trained on these limited labeled datasets. For longitudinal datasets, in [21] they proposed predicting the time difference between inputs from the two branches of a Siamese network as a pretext SSL method, the weights are then transferred for AMD progression modeling using 2D retinal OCT B-scans. Other popular pretext-based medical SSL methods exploit anatomies [22] or disease related biomarkers [4]. Later on, following the success of contrastive pretraining methods, in [23], they showed that when creating a positive pair for contrastive learning in a medical imaging setting, it was beneficial to use two different images from the same patient. Based on this idea, in [24], they proposed to generate pairs by sampling different frames from an ultrasound video, then the contrastive loss was used as a regularization term. In OCT domain, [6] used two images of a patient acquired at different dates for pretraining with a novel non-contrastive loss. Similarly, feeding two different images was used in stereo depth estimation for surgery videos [25]. Additionally, contrastive learning was applied to the unsupervised multimodal MRI registration task as a representation learning approach [26].

Several works have attempted predicting the AMD progression from retinal OCT. In [27], an LSTM-based model was trained to predict the conversion to dry-AMD within a specified time frame using a set of quantitative OCT biomarkers and demographic features. In [28], they boosted the performance of predicting the conversion to wet-AMD by exploiting the tissue segmentation maps and available disease diagnosis as auxiliary loss terms with 3D OCT volumes. In [29], they addressed predicting wet-AMD progression and obtained volume-level prediction by pooling the 2D B-scan-level predictions. [30] trained a CNN model using color fundus images and genotypes to predict late-stage AMD conversion over the long term (more than 2 years). The above works were largely trained from scratch, or from ImageNet-pretrained models, and suffered from the limited amount of labeled conversions available. Moreover, they rely on additional information such as patient demographics, supplementary disease diagnosis or generated segmentation maps, which are not always available or are extremely costly to obtain. In this regard, there is a need for exploiting large amount of unlabeled longitudinal OCT data available and find an effective self-supervised pretraining for the downstream predictive task.

Existing contrastive methods focus only on learning invariances to undesired image perturbations specific to 2D natural images. 3DTINC extends the invariances to 3D OCT volumes, and more importantly exploits the acquisition time information available in the temporal OCT datasets to learn representations that capture the longitudinal disease progression. After pretraining, 3DTINC performed better than baseline contrastive methods in the disease conversion prediction task.

## III. METHODS

Our approach consists of representation learning with the proposed 3DTINC, a non-contrastive pretraining on the longitudinal OCT volumetric data, followed by fine-tuning for predicting the conversion to wet-AMD (Fig. 2) within a six-month timeframe. The backbone for these steps is a compact and memory-efficient CNN designed for processing slice-based volumetric data.

### A. Model Backbone

The standard technique to expand a CNN from 2D to 3D is to inflate the 2D convolutional layers given that the input data is an isotropic 3D image of voxels. In an OCT volume, the resolution across the B-scan dimension is substantially lower than across the other ones. Thus, it is more meaningful to consider an OCT volume as a stack of slices (B-scans) similar to video frames, and use CNNs with smaller 3D kernels in the B-scan dimension to obtain a more efficient model [31], [32].

We chose *Channel-Separated Convolutional Network* (CSN) [32] as our backbone. CSN is more efficient compared to fully 3D CNNs or to common 3D video models in terms of number of parameters and flops. Its efficiency comes from combining $1 \times 1 \times 1$ 3D convolutions to process channel-wise information with $k \times k \times k$ depth-wise separable convolutions [33]. A 50-layer version of the CSN model is used for both non-contrastive pretraining and training the predictive model for the downstream task.

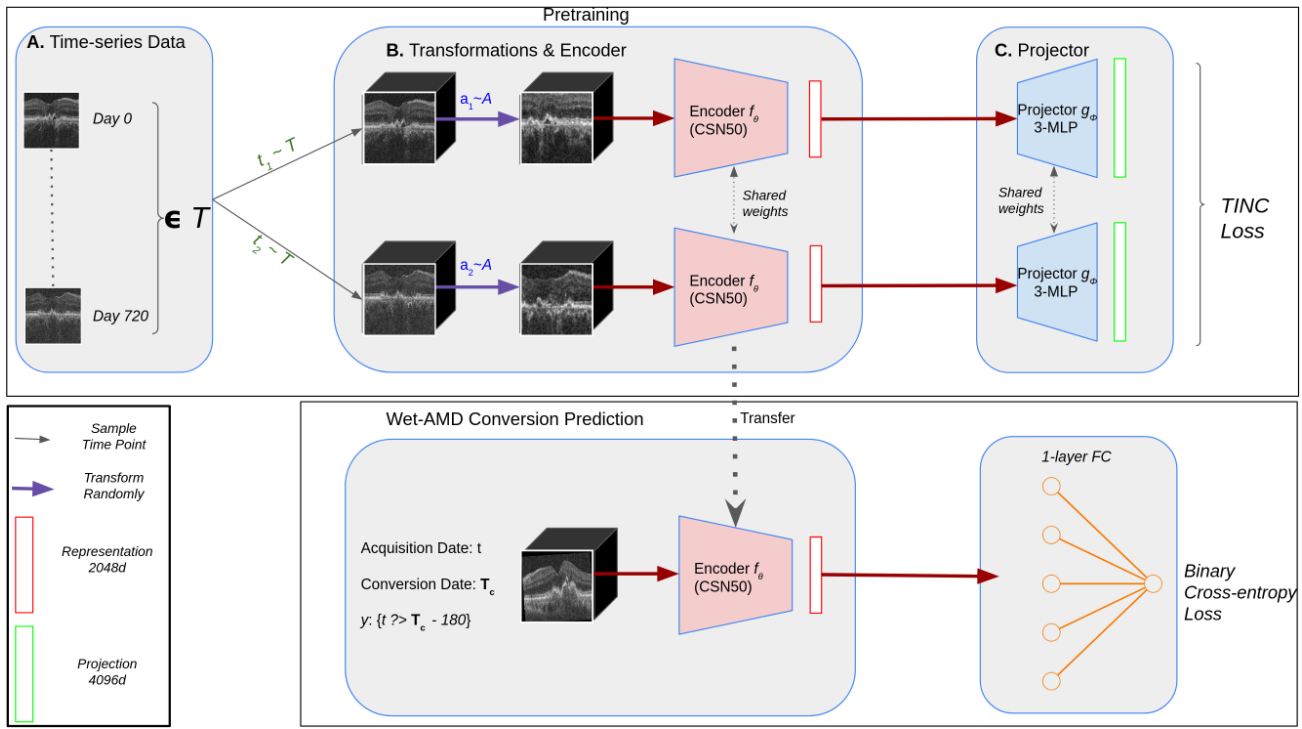

Fig. 2: Workflow of the proposed method. In the pretraining stage, two different visits from two time points $(t_1, t_2)$ of a patient, are randomly sampled. They are then transformed using 2 randomly sampled transformations $(a_1, a_2)$ to be fed to an encoder $f$ for extracting representations. The final non-contrastive loss *TINC* is calculated on the projections (*embeddings* of the representations). Finally, the weights of the pretrained encoder $f$ are transferred for learning the downstream task of wet-AMD conversion prediction, where binary cross-entropy loss is used for fine-tuning or linear evaluation.

### B. Proposed Non-Contrastive Pretraining

The proposed non-contrastive pre-training for disease progression has two novel aspects. First, the transformations used to create two views of the same scan for non-contrastive learning is specifically adapted for the retinal OCT modality and extended to operate on 3D Volumes. Secondly, we propose a novel time-aware similarity term in the non-contrastive loss which forces the network to learn similar representations for two 3D OCT scans of the same eye from two different time points, depending on the time interval between them. The OCT representations are extracted with two weight-sharing encoders (Fig. 2.B). Then, non-contrastive loss is calculated on the projections of the representations (Fig. 2.C)

*1) Proposed Volume Transformations:* Currently, image transformations for non-contrastive methods [5], [18] are based on the setting in [16], and are specific to the natural images. These transformations aim to capture the expected variability and perturbations in the images and the non-contrastive training aims to learn image representations which are invariant to them. Horizontal flip, color jittering, random cropping and blurring are some examples of the transformations (Table I). The contrastive learning creates two inputs (*views*) from an image by applying two transformations $a_1$ and $a_2$ sampled from a list of random augmentations $A$ (Fig. 2.B). One of the most important transformations in contrastive learning is the aggressive randomly resized crop [17]. In [5], [16], [18], the images are cropped between 8-100% of their areas. Although natural images could include some clues and information in very small crops, this is not the case with OCT images. The standardized view and the noisy nature of the OCT could result in an uninformative and extremely noisy crops. In this respect, medical OCT volumes require a distinct set of transformations, which the representations should be invariant to. Thus, we selected the cropping ratio to be between 40-80% of B-scans. We decreased the upper bound to 80% in order to mitigate the reduced augmentation strength by the higher lower bound [6]. In addition, we changed solarization threshold to $0.42$ to calibrate with respect to the intensity distribution of OCT volumes (Table I).

Inspired by [23], instead of creating two views from a single OCT, the input views are created from two OCTs of a patient acquired at different times acting as an additional transformation. This assumes that within a certain acquisition time interval, the change in anatomy is minimal, and the learned representations should encapsulate the patient-specific information. Finally, in order to learn representations invariant to spatial misalignment between the acquired intra-patient OCT volumes, we proposed shifting B-scan positions within an OCT volume as an additional transformation.

*2) Proposed Non-Contrastive Learning Similarity Loss:* Our non-contrastive pretraining approach is based on VICReg [5]. In comparison to other contrastive and non-contrastive methods, VICReg has some additional features that make it particularly well-suited for SSL on longitudinal volumetric data:
- There is no discrepancy between the branches of the

neural network. The weights of both encoders are updated simultaneously. This makes it easy to feed two different images to each branch.

- The loss terms are explicit and can be adapted easily. This is in contrast to methods like SimSiam [15], and BYOL [16], which do not have explicit loss terms or do not update the weights of both encoders simultaneously.
- Unlike Barlow Twins [18], the main similarity term does not require calculation of the cross-correlation matrix, which makes VICReg more robust to small batch sizes.

The loss is calculated using pair of projections $z$, and $z'$. The pretraining pipeline starts with transforming a volume $V$ twice via random transformations $a_1$ and $a_2$ to create two *views* as inputs to the Siamese network. Then the inputs are encoded by an encoder $f$ with learnable parameters $\theta$ (Fig. 2.B). Finally, the representations are expanded using projector [16] $g$ with parameters $\phi$ resulting in un-normalized projections $z = g_\phi(f_\theta(x))$ and $z' = g_\phi(f_\theta(x'))$ which the loss is calculated on (Fig. 2.C). Their batch versions are $Z = [z_1, \ldots, z_n]$ and $Z' = [z'_1, \ldots, z'_n]$ where $n$ is the batch size. A non-linear projector is helpful for preventing dimensional collapse [20], [34] and improving the downstream task.

VICReg uses three different loss terms. The *invariance term* $S(Z, Z')$ in (1) brings representations of two views of an image closer by decreasing the $L_2$ distance between $Z'$ and $Z$. The *variance term* $V(Z)$ in (2) ensures that there is a variation among the representations within a batch. It plays a similar role of pushing the negative samples away as in contrastive learning. The term itself is the sole factor preventing representational collapse, where the network learns to minimize the invariance loss by producing the same representation for all the inputs. In VICReg, it is implemented as a standard deviation instead of a direct calculation of variance to have more stable gradients [5]. In (2), $\gamma$ is the desired standard deviation (std) between all samples in a batch at each dimension of the projections, such that each projection in a batch, is different than the others to at least a degree of std, and similar to [5] the $\gamma$ is selected as 1. In (2), $\text{std}(z^j, \epsilon)$ is the standard deviation of $j$th dimension of the embeddings along a batch, a small $\epsilon$ is used for numerical stability. $d$ is the embedding space length. Lastly, inspired by the in-diagonal loss in [18], the *covariance term* $C$ ensures that each element of the embeddings are as informative as possible. The covariance matrix $Cov(Z)$ (3) is constructed by calculating the variance between each pair of projection vector elements. The loss term tries to bring off-diagonals of the $Cov(Z)$ close to zero so that each element in the vector is as informative as possible. The normalization could interfere with the decorrelation of the feature dimensions, (the covariance term) [34], hence the projections are un-normalized. In [5], they showed that $l2$ norm leads to degraded performance for VICReg.

$$S(Z, Z') = \frac{1}{n} \sum_i \|z_i - z'_i\|_2^2, \quad (1)$$

$$V(Z) = \frac{1}{d} \sum_{j=1}^{d} max(0, \gamma - \text{std}(z^j, \epsilon)), \quad (2)$$

$$C(Z) = \frac{1}{d} \sum_{i \neq j} [Cov(Z)]_{i,j}^2, \quad (3)$$

where $Cov(Z) = \frac{1}{n-1} \sum_{i=1}^{n} (z_i - \bar{z})(z_i - \bar{z})^T$

**Temporally informed non-contrastive (TINC) loss:** In Section III-B.1, we proposed creating two views not from a single image but with scans from two visits of a patient. Given two different visits at time points $t_1$ and $t_2$, $\Delta t$ is defined as the time interval $|t_1 - t_2|$ normalized to the range of $[0, 1]$. We hypothesize that as the time difference $\Delta t$ increases, the distance between $z$ and $z'$ should be within a margin proportional to $\Delta t$. In the longitudinal OCT data, the increase in anatomical changes over time should be expected due to disease progression. Since the VICReg invariance term $S(Z, Z')$ in (1) does not capture this change, we propose to replace it with a novel margin-based invariance (i.e. similarity) term, $\ell_{\text{TINC}}$ [6] defined as:

$$\ell_{\text{TINC}}(Z, Z') = \frac{1}{n} \sum_i^n max(0, \|z_i - z'_i\|_2^2 - \Delta t_i) \quad (4)$$

$\ell_{\text{TINC}}$ is a generalization of the VICReg invariance loss term designed to make the learned representations sensitive to time by using an adaptive margin. The margin is determined by the $\Delta t$, and dynamically changed for each visit pair. It brings $z$ and $z'$ close only if the distance between them is greater than $\Delta t$. As the margin approaches 0, $\ell_{\text{TINC}}$ becomes the invariance term of VICReg.

$\ell_{\text{TINC}}$ is similar to the $\epsilon$-insensitive loss [35] used in support vector regression (SVR) with some key differences. $\ell_{\text{TINC}}$ uses an adaptive margin dependent on $\Delta t$ instead of a constant margin $\epsilon$ in SVR. Additionally, our margin is between 0 and $\Delta t$, not $-\Delta t$ and $\Delta t$, because the distance metric cannot provide negative values.

For the contrastive training, we kept variance and covariance terms as they are in VICReg, and the final loss function is the weighted sum of these three terms:

$$L_{\text{contr}}(Z, Z') = \lambda \cdot \ell_{\text{TINC}}(Z, Z') + \mu \cdot (V(Z) + V(Z')) \quad (5)$$
$$+ \nu \cdot (C(Z) + C(Z'))$$

Although the variance term in (2) acts to increase the standard deviation in each feature dimension across scans of different eyes in a mini-batch, it also encourages a larger vector norm, which in turn increases the distance between $z$ and $z'$ implicitly. The trade-off between the $\ell_{\text{TINC}}$ and the variance term plays a crucial role in learning meaningful representations. Ideally, when the distance between $z$ and $z'$ is above the margin, the $\ell_{\text{TINC}}$ term should decrease the distance to bring it closer to the margin. However, in a scenario when the initial distance is below the margin during training, the $\ell_{\text{TINC}}$ does not contribute to the overall loss and the variance term implicitly moves $z$ and $z'$ by the encouraging them to This article has been accepted for publication in IEEE Transactions on Medical Imaging. This is the author's version which has not been fully edited and content may change prior to final publication. Citation information: DOI 10.1109/TMI.2024.3391215

TABLE I: Contrastive Learning Transformations

| Augmentation | Original Setting [16] | Proposed Setting |
|---|---|---|
| Random Crop & Resize (percentage) | 0.08 - 1.0 | **0.4 - 0.8** |
| Random Horizontal Flip (probability) | 0.5 | 0.5 |
| Random Color Jittering (probability) | 0.8 | 0.8 |
| Random Gaussian Blur (kernel size) | 21 | 21 |
| Random Solarize (threshold) | 0.5 | **0.42** |
| Random Slice Shift (frame) | - | **±5** |
| Input Time Difference (days) | - | **90-540** |

*Random grayscale transformation is removed for all cases, since the B-scans are already in grayscale.

have high vector norms. Thus, at the convergence point, the loss function encourages the distance between $z$ and $z'$ to be close to $\Delta t$ for a given input pair. During training, two different visits of a patient are randomly selected in each epoch to form the input image pairs. This ensures a wide variation in the $\Delta t$ for the training samples corresponding to each patient in every epoch. It is crucial that a patient contributes only a single pair to a mini-batch in order to prevent treating the different scans from a patient as implicit negatives.

*3) Temporal equivariance property:* The goal of contrastive learning is to have representations that are *invariant* to the expected irrelevant image distortions. However, disease-related changes in the anatomy over time should have a proportional effect on the representational space. Thus, it is desirable to have *equivariant* representations across sufficient time difference. Formally, suppose the distance between the representations of two intra-patient OCTs, $x_t$ and $x_{t+\Delta t}$ encoded by encoder $f$, with time difference of $\Delta t$. Now, if $t + \Delta t$ is increased by $\Delta t$, Euclidean distance between $f(x_{t+\Delta t})$ and $f(x_{t+2\Delta t})$ should be similar to the distance between $f(x_t)$ and $f(x_{t+\Delta t})$. If $\Delta t$ is small (one month), the change in distance between $x_t$ and $x_{t+\Delta t}$ can be represented by a small transformation $h$. This assumes that with small $\Delta t$, the transformation between two consecutive scans does not cause label-related semantic changes. Thus, if $f$ is producing time-equivariant representations, following the formulation in [36] this property can be given as:

$$f(x_{t+2\Delta t}) \approx hf(x_{t+\Delta t}) \approx h \circ hf(x_t). \tag{6}$$

If $\Delta t$ induces a small temporal neighborhood, $h$ can be approximated using its first-order Taylor series as:

$$f(x_{t+\Delta t}) \approx f(x_t) + \Delta t c(t), f(x_{t+2\Delta t}) \approx f(x_t) + 2\Delta t c(t) \tag{7}$$

Additionally if the set of possible values for $t$ is bounded by a relatively close upper and lower bounds, the derivative with respect to time can be approximated by a constant function, which is $c(t)$. If we take the $l2$ distance between two time points:

$$\|f(x_{t+\Delta t}) - f(x_t)\|_2 \approx \|f(x_t) + \Delta t c(t) - f(x_t)\|_2 \tag{8}$$
$$\approx \|\Delta t c(t)\|_2$$

Similarly $\|f(x_{t+2\Delta t}) - f(x_t)\|_2$ will be $2\|\Delta t c(t)\|_2$ and $\|f(x_{t+2\Delta t}) - f(x_{t+\Delta t})\|_2$ will be $\|\Delta t c(t)\|_2$. In other words, the distance between two time-equivariant features is a function of the scalar $\Delta t$. $\ell_{\text{TINC}}$ tries to achieve a similar result;

the distance between two representations of the time points should be 0 except for the *normalized time difference*. It is important to note that we formulated $\Delta tc(t)$ as $\Delta t$ normalized between 0-1 depending on the maximum and minimum values of $t$. Instead of enforcing Eq. (7) as in [36], $\ell_{\text{TINC}}$ addresses the time equivariance implicitly. Additionally, in the case of $\ell_{\text{TINC}}$, if the other loss terms are ignored during training, the trivial solution would still produce the representations specific to each time difference $\Delta t$, unlike in [5], [18], where the trivial solution produces the same representation for all inputs.

### C. Downstream Progression Task Formulation

We formulated the intermediate AMD to wet-AMD stage progression task as a binary classification of predicting the conversion event within a fixed time period from a given patient visit. Clinically, it is desirable to predict the conversion to late stage wet-AMD within a short time interval to start the treatment with anti-VEGF drugs in a timely manner. Furthermore, it is important to be able to make the prediction from a single visit (i.e., a single OCT scan), so that the patient's condition can also be assessed from the first, baseline visit, and not all the datasets have regular visit intervals and complete scans. Similarly to the other deep learning works addressing the same task [21], [27]–[30], we chose the time period as 6 months to be considered clinically relevant. The pretrained model is fine-tuned end-to-end for this downstream prediction task.

The binary conversion labels are hence assigned to each scan. A scan, sampled from a longitudinal study of $T$ months, is considered a member of a positive class if there is a detected wet-AMD conversion point within the next 6 months; otherwise, it is assigned to the negative class. Following this reasoning, there is no possible label assignment for scans of the already converted eyes; therefore, those scans were excluded from the dataset. For a 3D OCT volume $V_c$, where $c$ is the first visit date of wet-AMD diagnosis in months for a patient $P$, the label $y_t$ associated with volume $V_t$ at visit month $t$ is generated as:

$$t \in \{x \in \mathbb{Z} \mid T \geq x \geq 0\}, y_t = \begin{cases} 0 & if\ c - t > 6 \\ 1 & if\ 0 \leq c - t \leq 6 \\ No\ label & if\ c - t < 0 \end{cases} \tag{9}$$

## IV. EXPERIMENTAL SETUP

All experiments and models are implemented using PyTorch with mixed precision floating point and MONAI [37], including the linear evaluation step. The pretraining, fine-tuning and linear evaluation steps were performed on a single 80 GB, NVIDIA A100 GPU.

### A. Datasets

The HARBOR clinical trial dataset [1] is utilized to train and evaluate the proposed method. The trial examined the efficacy of anti-VEGF treatment on wet AMD eyes over a 24-month period. It consists of monthly visits where both eyes of each subject were imaged with Cirrus OCT (Zeiss Medictec, US) with 128 B-scans, each with a size of $1024 \times 512$ px, covering $2mm \times 6mm$. From this dataset, we extracted the OCTs of the fellow eyes, which were at the intermediate AMD stage at baseline. We identified 463 such fellow eyes that were labeled by two retinal experts [38], out of which 117 converted to wet-AMD during the trial duration, and 346 remained in the intermediate stage. In total, there are 12506 OCT scans for the SSL and 10108 for the downstream task (Table II).

---

[1]NCT00891735. https://clinicaltrials.gov/ct2/show/NCT00891735

TABLE II: Details of Self-Supervised Learning and Downstream Supervised Learning Tasks.

|  | Self-Supervised | Supervised | Supervised (External) |
|---|---|---|---|
| Dataset | HARBOR | HARBOR | PINNACLE |
| OCT scanner | CIRRUS | CIRRUS | TOPCON |
| #Patients | 540 | 463 | 334 |
| Mean Scan Interval [Days] | 30 | 30 | 126 |
| #Scans | 12506 | 10108 | 2813 |
| #Converter eyes | n/a | 117 | 127 |
| #Converter scans (within six months) | n/a | 547 | 536 |

In order to evaluate the cross-domain adaptability of the representations extracted by the models pretrained on HARBOR dataset, we used an additional dataset from the PINNACLE consortium [39]. It consists of 2813 OCT scans from 127 converter and 207 non-converter eyes (Table II). Patient visits were irregular with a mean interval of 126 days between 2 consecutive acquisitions. The eyes were scanned with the Topcon scanner, which produces OCT volumes with 128 B-scans with a size of $885 \times 512$ px, covering $2mm \times 6mm$. We note that a difference in OCT scanners between the two datasets introduces an image domain-shift.

### B. Preprocessing

We used the central 32 B-scans covering $2 \times 3 \times 6\ mm^3$ ($axial \times B-scans \times A-scans$) volume of the retinal tissue around the macula as input to our 3D network during both the pretraining and the supervised training steps. The B-scans in this region contain the most relevant clinical information for AMD and also reduce the GPU memory requirement. The curvature of the retinal tissue was flattened with respect to the Bruch's Membrane which was automatically segmented using the method in [40]. Flattening is commonly employed in deep learning applications to have a standardized retinal shape and reduce misalignments between the B-scans of an OCT volume. Next, the axial extent of each B-scan was cropped by removing the dark background regions above and below the retinal tissue to obtain a $0.78 \times 3 \times 6\ mm^3$ region of interest which is resized to $32 \times 224 \times 224$ voxels. The voxel intensities are normalized between 0-1 using the min-max scaling. The preprocessing steps are the same for both datasets.

### C. Experiments

*1) Pretraining:* In order to create input pairs from two different time points, the method should not require any discrepancy between the branches of the contrastive model. We hence chose VICReg and Barlow Twins as baselines because from all the current non-contrastive methods, they are the only ones that do not rely on asymmetry between the encoders. To establish a fair comparison, we first trained the original VICReg and Barlow Twins. Second, we applied our proposed 3D OCT transformations (Section III-B.1) on VICReg and Barlow Twins to obtain their *improved* versions. Thus, the original VICReg and the improved VICReg serve as ablation studies to our proposed 3DTINC. We also pretrained a common contrastive method BYOL [16] and tested it in our downstream datasets. Additionally, we compared 3DTINC against another equivariant contrastive learning called EquiMod [41]. Equimod has two distinct projections of the representation space. One projection is dedicated to addressing the invariance task, while the second one is designed for the equivariance task. Within the equivariance branch, a specialized module is trained to predict the transformed representation, using the untransformed representation and the image transformation parameter (time difference).

In order to show the benefits of pretraining, for comparison, we initialized a CSN model with random weights, and completed a fully supervised training on it. Additionally, we fine-tuned a common video prediction architecture SlowFast [31] with pretrained weights learned from Kinetics dataset [42] to demonstrate the importance of pretraining using the same medical imaging modality as the downstream task.

*2) Downstream Task:* We evaluated the learned representations on the downstream task using linear evaluation and fine-tuning. We split our supervised data for the downstream prognostic task into a development set (80%) and a hold-out test set (20%). The development set was used for a four-fold patient-wise cross-validation to train four models and to tune the hyperparameters. The folds were stratified by the binary conversion label. The four models were then evaluated on the holdout test set, where their mean and standard deviation of the performance were evaluated.

### D. Implementation Details

*1) Pretraining:* When pretraining, only a small batch size of 32 could be used for pretraining the Barlow Twins and VICReg models, which makes them very sensitive to hyperparameters. We pretrained CSN using vanilla Barlow Twins with a learning rate of 0.005 with a cosine decay and a weight decay of 0.001. In Barlow Twins training, the weight of the off-diagonal loss term is changed to 0.01. For vanilla VICReg training, we change the learning rate to 0.0001 and the weight decay to 1e-6. Weight decay is removed for biases and BatchNorm statistics. We kept the loss terms weights the same as described in [5]. For all pretraining experiments, we employed AdamW optimizer [43]. In improved versions of Barlow Twins and VICReg, the learning rates are set to 5e-5 and 5e-4 respectively. In the 3DTINC experiments, learning rate of 5e-4 and weight decay of 1e-4 are used. The weights of the 3DTINC loss terms are set as $\lambda=15$, $\mu=25$ and $\nu=5$ based on the convergence rate of the individual loss terms. We pretrained all models for 400 epochs (10 warm-ups). All models are pretrained with the same number of optimization steps. In vanilla settings, an epoch length is (#images / Batchsize), while in 3DTINC it is (#patients / Batchsize). In order to make the total number of pretraining steps equal, during 3DTINC pretraining, we resampled multiple time-dependent OCT pairs per patients. For the improved Barlow Twins, VICReg and 3DTINC, in order to ensure that the anatomical changes are present but still representational similarity between two time steps can be established, we bounded it between 90 and 540 days. The rationale for the upper and lower bounds is to have a better time-equivariance property, as explained in Section III-B.3. It is important to note that we made sure that there is only one pair per patient in a batch, in order to prevent contrastive effect stemming from the loss terms.

During pretraining, feature representations from the encoder are first projected onto a high-dimensional space using the projection head $g_\phi$ as described in Section III-A. The non-contrastive loss is calculated on the projections. $g_\phi$ is implemented as a multi-layer perceptron (MLP) with two hidden layers, each having the dimension of 4096 with BatchNorm and ReLU activation. Finally, the CSN and projector $g_\phi$ have $\sim 11M$ trainable parameters.

*2) Downstream Task:* For linear evaluation, we discarded the MLP projector and added a single linear layer after the representations calculated by $f_\theta(x)$ [11]. The linear layer is trained for 50 epochs on the conversion task using Adam optimizer [44] with a constant learning rate $10^{-4}$. When fine-tuning, a fully connected layer is attached after the representations to get the logits. During fine-tuning, the model with pretrained weights is updated end-to-end with a learning rate $10^{-4}$ using the Adam optimizer for 100 epochs. When training the model from randomly initialized weights, the model is trained for 300 epochs using AdamW with an initial learning rate of $5 \times 10^{-4}$ decayed by a cosine scheduler and weight decay of $10^{-6}$. The best epoch with the highest ROCAUC score is chosen on the validation set during fine-tuning and training from random weights. The classification loss (binary cross-entropy loss) in all the downstream experiments was weighted $5:1$ in favor of the positive class to mitigate the heavy class imbalance. In all downstream task settings, we applied random translation, small random rotation (15 degrees) and random horizontal flip as augmentations.

*3) Metrics:* Because the scan labels are heavily imbalanced towards the negative class ($\approx 1:20$), we reported the results in terms of the area under the receiver operating characteristic (ROCAUC), the area under the precision-recall curve (PRAUC), and the balanced accuracy (BACC) calculated as the average recall of the classes. In the HARBOR dataset, the baseline PRAUC value is 0.054, while in the PINNACLE dataset, the PRAUC baseline is 0.097. The low baseline values of PRAUC are due to the large imbalance within the datasets. ROCAUC is chosen as the primary metric to compare performance between models. The reason is that we are more interested in the correct ranking of classes and we are not applying a detailed fine-tuning scheme in the downstream task for better prediction probabilities, which can lead to many False Positives.

## V. RESULTS

### A. Linear Evaluation of Learned Representations

The quantitative results of the linear evaluation on the two datasets are shown in Table III. We first tested models pretrained with Barlow Twins & VICReg by extending the network architecture and the original augmentations to 3D. The increase in the sizes of the model and the input affected the maximum batch size that fits into the memory. Compared to [6], Barlow Twins performed worse than VICReg, achieving only $0.618$ ROCAUC. It is due to the limited batch size causing numerical instabilities during the calculation of cross-covariance matrix for Barlow Twins loss. Then, in order to investigate the proposed OCT specific transformations (Table I), we pretrained Barlow Twins and VICReg with them. Both Barlow Twins and VICReg had higher prediction scores once the pretrainings were reinforced with the novel augmentations and the input scheme. This demonstrates the importance of having an OCT specific transformation for contrastive pretraining. When we replaced the similarity loss with $\ell_{\text{TINC}}$ in (4), 3DTINC pretraining achieved the highest score with $0.792$ ROCAUC compared to the other pretraining settings. We provide a sample of saliency maps from the 3DTINC linear evaluation in Fig. 3. Additionally, EquiMod achieved a comparatively high ROCAUC score of $0.785$, demonstrating the importance of the temporal equivariance property. In terms of ablation studies, OCT-specific augmentations and the two-time-point input improved the original VICReg.

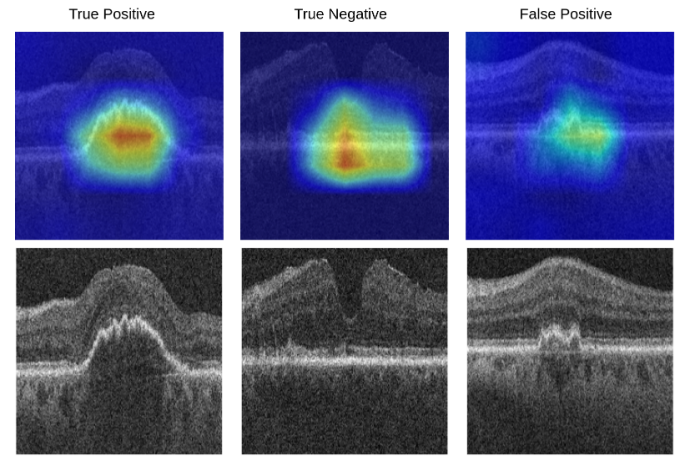

Fig. 3: Examples of Grad-CAM saliency maps from the linear evaluation predictions.

### B. Robustness to Image Domain Shift

OCT scanners from different vendors use different approaches and protocols during image acquisition. These inherent differences are sufficient to cause image domain shift between different scanners [45]. We compared the multiplicative speckle noise characteristics of the PINNACLE and the HARBOR datasets with the Gray Level Co-occurrence Matrix (GLCM) contrast value. We obtained an average contrast value of 182.4 for the HARBOR dataset and 138.5 for the PINNACLE dataset, indicating a strong difference in noise

TABLE III: *Linear Evaluation* Results for Wet-AMD Conversion Prediction within 6 Months.

| Method | HARBOR | | | External Data: PINNACLE | | |
|---|---|---|---|---|---|---|
| | ROCAUC ↑ | PRAUC ↑ | BAcc ↑ | ROCAUC ↑ | PRAUC ↑ | BAcc ↑ |
| BYOL (expanded to 3D) | 0.709 ± 0.016 | **0.176 ± 0.020** | 0.612 ± 0.029 | 0.575 ± 0.020 | 0.122 ± 0.010 | 0.549 ± 0.021 |
| Barlow Twins (expanded to 3D) | 0.618 ± 0.007 | 0.155 ± 0.005 | 0.564 ± 0.011 | 0.593 ± 0.019 | 0.121 ± 0.007 | 0.556 ± 0.034 |
| Barlow Twins (improved) | 0.771 ± 0.007 | 0.156 ± 0.014 | 0.708 ± 0.008 | 0.607 ± 0.013 | 0.125 ± 0.010 | 0.564 ± 0.023 |
| VICReg (expanded to 3D)* | 0.724 ± 0.024 | 0.136 ± 0.009 | 0.616 ± 0.019 | 0.601 ± 0.011 | 0.152 ± 0.016 | 0.557 ± 0.027 |
| VICReg (improved)* | 0.757 ± 0.017 | 0.144 ± 0.014 | 0.612 ± 0.017 | 0.613 ± 0.014 | 0.172 ± 0.015 | 0.546 ± 0.021 |
| EquiMod | 0.785 ± 0.016 | 0.157 ± 0.009 | 0.621 ± 0.007 | 0.610 ± 0.014 | 0.126 ± 0.004 | 0.574 ± 0.039 |
| **3DTINC** | **0.792 ± 0.018** | 0.159 ± 0.010 | **0.708 ± 0.007** | **0.675 ± 0.010** | **0.229 ± 0.008** | **0.603 ± 0.028** |

* denotes the ablation studies.

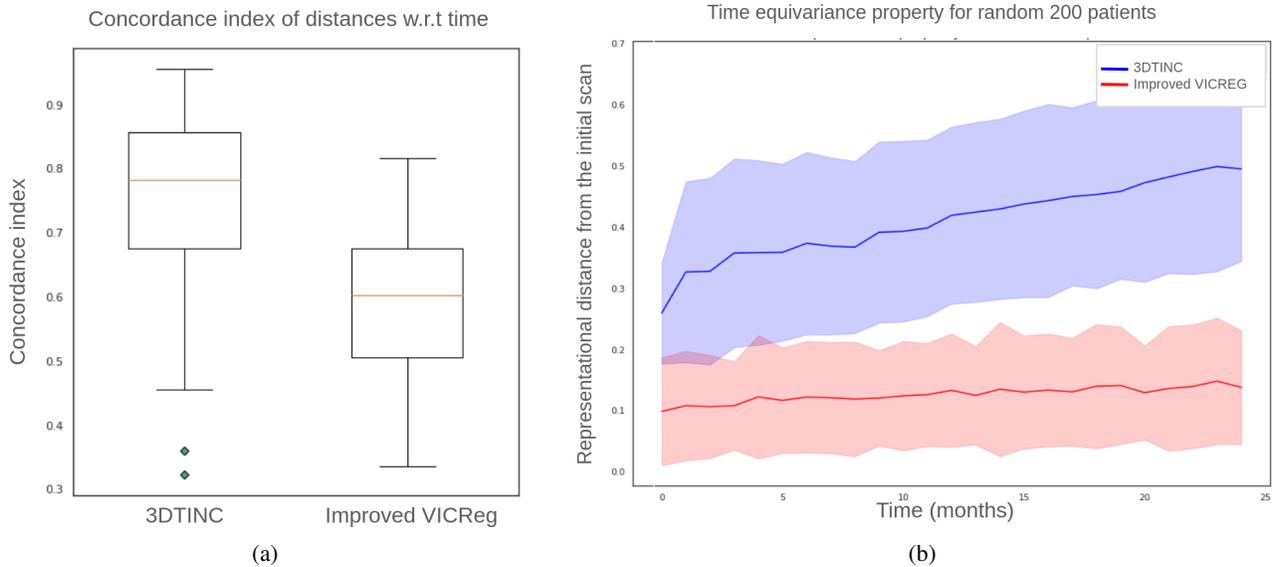

Fig. 4: Population level (random 200 patients) analysis of the time-equivariance property. **(a)** Concordance index on the ranking of distances. **(b)** Average representational distance between a given month and the initial month for a patient.

TABLE IV: *End-to-end Fine-tuning* Results for Wet-AMD Conversion Prediction within 6 Months.

| Method | HARBOR | |
|---|---|---|
| | ROCAUC ↑ | PRAUC ↑ |
| CSN trained from scratch | 0.722 ± 0.026 | 0.150 ± 0.032 |
| SlowFast pretrained on Kinetics | 0.753 ± 0.005 | **0.194 ± 0.007** |
| BYOL (expanded to 3D) | 0.751 ± 0.020 | 0.162 ± 0.015 |
| Barlow Twins (expanded to 3D) | 0.649 ± 0.022 | 0.141 ± 0.013 |
| Barlow Twins (improved) | 0.772 ± 0.007 | 0.148 ± 0.015 |
| VICReg (expanded to 3D)* | 0.763 ± 0.011 | 0.140 ± 0.018 |
| VICReg (improved)* | 0.775 ± 0.013 | 0.170 ± 0.013 |
| EquiMod | 0.730 ± 0.044 | 0.154 ± 0.027 |
| **3DTINC** | **0.781 ± 0.002** | 0.162 ± 0.004 |

* denotes the ablation studies.

(Fig. 5). In order to show the robustness of the extracted OCT representations, we compared different pretrained models on an external dataset acquired with a different OCT scanner (Table II). The linear evaluation results showed that 3DTINC is able to produce more linearly separable representation even under domain shift by achieving 0.675 ROCAUC. On the other hand, the other models, including improved versions of VICReg and Barlow Twins, performed close to random (0.5 ROCAUC score) or slightly above it (Table III). Although we did not enforce domain invariance during pretraining, the results showed that better pretraining on the source data could lead to more domain-invariant representations compared to the other pretraining approaches. However, the high ROCAUC score of 3DTINC (Table III) underscores its ability to learn better features compared to other methods, which would be advantageous even under domain shift. Even though 3DTINC results in better performance under the domain shift, it is clear that there is a big gap for achieving domain-invariance between different OCT scanners. We leave it as a future work to optimize for the domain invariance in the pretraining step.

### C. Fine-Tuning on the Downstream Task

When the pretrained models were fine-tuned in an end-to-end fashion, 3DTINC outperformed the other models with more confidence (standard deviation of 0.002 for ROCAUC Table IV). All other pretraining models achieved higher RO-CAUC scores against the baselines, except 3D Barlow Twins with the original setting due to the batch size related problems. Additionally, when the 3D models were compared with their 2D counterparts in [6], it showed that 3D information allowed the network to capture more details for the downstream task. Also, a fine-tuned SlowFast achieved better performance than original VICReg and Barlow Twins, but performed worse once they were improved. Finally, we observed very quick overfitting in fine-tuning even with hyperparameter tuning on the validation set. The overfitting resulted in a degraded performance for fine-tuning compared to linear evaluation.

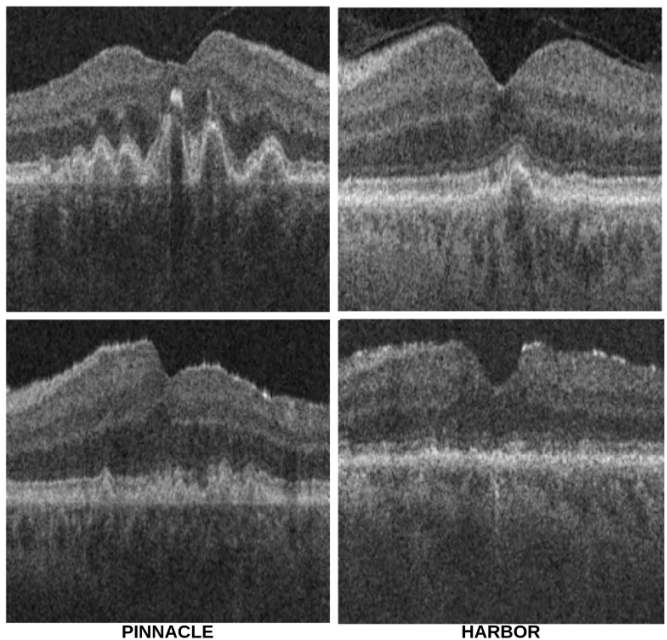

Fig. 5: An example of 2 B-Scans from the HARBOR and the PINNACLE datasets to demonstrate the difference in the noise characteristics.

### D. Time-Equivariance Analysis

One of the important properties of the proposed method is that the loss function converges to provide time-equivariant features (Section III-B.3). We hypothesised that in order to extract time-equivariant representations, the network should be aware of anatomical biomarkers that change over time, which in turn facilitates learning the downstream progression detection task. Unlike the other equivariance methods where the transformation parameters are estimated directly [10], [21], the TINC loss (4) enforces it with the margin in the loss function as a soft constraint. To demonstrate the equivariance property, we sampled 200 patients randomly, and calculated the $l_2$ distance between each visit and the patient's initial visit in the representational space (Fig. 4b). We compared 3DTINC against improved VICReg where both networks have inputs from two different visits. We observed that there is a clear increasing trend in distance as the time progresses from the initial visit. Although the model only exposed to the time difference between 3-18 months, the graph clearly demonstrates that the network is clearly aware of time regardless of the time difference from the initial scan. In contrast to 3DTINC, the representations obtained from improved VICReg, do not show a clear increase in distance with respect to time, even though the model improved with inputs from different time steps. In order to show that the difference of the slopes in Fig 4b is not due to the scale, we calculated the concordance index (CI) averaged over 200 patients to measure whether the distances rank with respect to time or not (Fig 4a). A higher CI value of 3DTINC shows that the distances with respect to the initial scan increase as time progresses. The mean CI is close to 0.5 for the improved VICReg indicating an almost random ranking. In summary, we believe that the time-equivariance property is crucial for settings where the prediction is performed from a single scan. This is because if the network is able to produce time-equivariant representations, then future OCT scan representations can be predicted from the current visit's representation and the desired future time intervals without seeing the actual future scans.

## VI. CONCLUSION

Predicting the future risk of conversion from the intermediate stage of AMD to its advanced wet stage with prognostic deep learning models can play a critical role in individualized patient management by enabling frequent monitoring and timely treatment for patients with a faster rate of disease progression. Such prognostic models require labeled longitudinal imaging datasets, which are costly to obtain and feature high class imbalance. To exploit large amounts of unlabeled longitudinal volumetric data available in eye clinics, we adapted non-contrastive SSL methods as a pretraining step, which have the memory efficiency required for learning from 3D OCT volumes. Importantly, our novel similarity loss term for longitudinal pretraining enables the network to account for temporal changes in the retinal anatomy by encouraging the learned representations to be time equivariant. The results of our large-scale internal and external evaluation showed that such pretrained models are more successful on the downstream task of predicting conversion from intermediate to wet-AMD. The fine-tuned model provides a prognosis from a single OCT volume without requiring multiple past visits, thereby allowing risk assessment from the patient's first visit itself.

Although to a lesser extent than contrastive learning, non-contrastive learning methods still employ some loss terms which are sensitive to the training batch size. Thus, even though we demonstrated that non-contrastive SSL pretraining improved downstream performance, there remains room for improvement by pretraining with larger batch sizes, which is currently restricted by the large GPU memory requirements of 3D CNN architectures. Thus, exploration of more efficient 3D deep learning models, e.g., through neural architecture search, could be a direction for future work. Another promising avenue for future work is to design prognostic models that consider multiple visits from the patient's history to predict the risk of conversion instead of a single baseline visit.